\documentclass[11pt]{article}

\usepackage[T1]{fontenc}
\usepackage[utf8]{inputenc}
\usepackage{lmodern}
\usepackage{microtype}
\usepackage[margin=1in]{geometry}
\usepackage{amsmath,amssymb}
\usepackage{graphicx}
\usepackage{booktabs}
\usepackage{longtable}
\usepackage{etoolbox}
\AtBeginEnvironment{longtable}{\small\setlength{\tabcolsep}{3pt}}
\usepackage{array}
\usepackage{tabularx}
\usepackage{pdflscape}
\usepackage{multirow}
\newcolumntype{Y}{>{\raggedright\arraybackslash}X}
\newcolumntype{C}{>{\centering\arraybackslash}X}
\usepackage{caption}
\usepackage{subcaption}
\usepackage{xcolor}
\usepackage{xurl}
\usepackage{hyperref}
\hypersetup{
  pdftitle={Factual Yet One-Sided: Stance Poisoning and Evidence Balancing in Retrieval-Augmented Generation}
}
\usepackage{cleveref}
\IfFileExists{natbib.sty}{\usepackage[numbers,sort&compress]{natbib}}{}

\providecommand{\tightlist}{%
  \setlength{\itemsep}{0pt}\setlength{\parskip}{0pt}}

\title{Factual Yet One-Sided: Stance Poisoning and Evidence Balancing in Retrieval-Augmented Generation}
\author{
  Puyuan Yang\textsuperscript{1,\(\ddagger\),\textsection} \and
  Shaoxuan Dong\textsuperscript{2,\(\ddagger\)} \and
  Liujie Wu\textsuperscript{1} \and
  Xiaochun Cheng\textsuperscript{1,*} \and
  Bo Li\textsuperscript{2,*}\\[0.75em]
  \small \textsuperscript{1}Department of Computer Science, Swansea University, Swansea, United Kingdom\\
  \small \textsuperscript{2}School of Electronics and Information, Northwestern Polytechnical University, Xi'an, China\\[0.5em]
  \small \textsuperscript{\(\ddagger\)}These authors contributed equally.\\
  \small \textsuperscript{\textsection}Email: \href{mailto:yangpuyuan123@gmail.com}{yangpuyuan123@gmail.com}\\
  \small \textsuperscript{*}Corresponding authors:
  \href{mailto:xiaochun.cheng@swansea.ac.uk}{xiaochun.cheng@swansea.ac.uk};
  \href{mailto:libo803@nwpu.edu.cn}{libo803@nwpu.edu.cn}
}
\date{}

\begin{document}

\maketitle

\begin{abstract}
The stance expressed in responses from retrieval-augmented generation
systems is significantly influenced by the stance distribution of the
retrieved materials. Existing research on knowledge poisoning has
primarily targeted factual errors. In contrast, this study focuses on
injection attacks in which the facts are correct but the stance on a
controversial issue is one-sided. At a mean topic-level poisoning share of
18.9\% within the post-injection ProCon documents, all four generation models from different vendors
exhibited stance shifts aligned with the poisoning direction; across
four concentration levels, the shifts increased monotonically.

This paper proposes a Stance-Balancing Middleware positioned between the
retriever and the generator. It consists of stance detection, fixed
balancing (2 supporting + 2 opposing + 1 neutral), and a debiasing
prompt, without introducing additional large language model calls. Across all eight concentration--direction settings, the complete method reduced the magnitude of stance deviation by an estimated 0.106 to 0.350; the six settings at poisoning shares of 11.9\%--18.9\% remained significant after Holm adjustment (adjusted p \(\leq\) 0.009). In the cross-model experiment, all eight model--direction combinations moved along the stance axis in the corrective direction, and seven of eight showed a clearly detected reduction in absolute distance from 0. In the concentration sweep, the post-defense intervals included 0 in all eight
settings, while the cross-model pro-poisoning results retained small positive
residual point estimates of +0.045 to +0.063. For 20 candidate documents, stance detection took approximately 1.5\% of the median generation time on a single GPU and approximately 16\% on CPU.


\end{abstract}

\section{Introduction}\label{sec:introduction}

\subsection{Background}\label{subsec:introduction-background}

Retrieval-Augmented Generation (RAG) connects an updatable external
knowledge base to a large language model. After a user submits a
question, the system first retrieves relevant documents from the
knowledge base and sends them to the model together with the question,
enabling the model to formulate its response on the basis of the
retrieved material. This paradigm mitigates problems arising from
outdated knowledge encoded in model parameters and fabricated content
and has been applied in fields including healthcare, finance, aviation
safety, industrial operations and maintenance, and education.

At the same time, this paradigm introduces a less-discussed property:
the stance of a model\textquotesingle s response can be influenced by
the retrieved material. The more faithfully the model follows the
retrieved material, the more likely the stance distribution of that
material is to carry over into the response. For factual questions, this
provides traceable support for the answer. For controversial questions,
however, the selection of materials affects which stances are presented.
Controlling the composition of the retrieval pool may therefore provide
a means of manipulating the stance conveyed by model outputs.

\subsection{Motivation and
Challenges}\label{subsec:introduction-motivation}

Most RAG-poisoning studies target factual corruption or attacker-specified
outputs. Recent work has also demonstrated topic-level opinion manipulation,
showing that retrieval can be exploited to steer model positions across related
queries. This paper focuses on a complementary question: how to defend against
factually grounded but stance-skewed document injection without modifying the
retriever or generator.

Specifically, we study source-grounded document injection that concentrates
evidence on one side of a controversial issue; human validation of the sampled
injected documents is reported in Section~\ref{sec:experimental-setup} and
Appendix~\ref{sec:human-document-validation}. Controversial issues generally have
no single widely accepted answer, and both sides possess genuine supporting
arguments. An attacker need not fabricate content; injecting arguments from one
side in bulk is sufficient to alter the stance composition of the retrieval
pool. Checking individual claims alone does not establish whether the retrieved
evidence represents competing positions adequately.

We constructed poisoned knowledge bases in two directions for 210
controversial questions. At a mean topic-level poisoning share of 18.9\%
within the post-injection ProCon documents (Section~\ref{sec:experimental-setup}),
all four generation models---from different vendors and with
varying parameter counts---exhibited stance shifts aligned with the
poisoning direction, and the shifts increased monotonically with
poisoning concentration (see Figs.~\ref{fig:attack-defense} and~\ref{fig:cross-model}).

The attack is difficult to detect because the affected responses may
remain factually valid: the cited materials can be verified, but they
present arguments from only one side. Evaluations based on answer
correctness struggle to capture this type of bias. The problem lies in
the composition of stances, a dimension for which existing evaluation
and defense systems lack adequate measurement. This study therefore
begins by establishing such a measure.

\subsection{Contributions}\label{subsec:introduction-contributions}

First, this study characterizes poisoning through stance skew rather than
factual error. We define a continuous stance score over sets of controversial
questions and use strictly paired effect sizes and bootstrap confidence
intervals to quantify how much a defense reduces deviation from 0.

Second, we propose a Stance-Balancing Middleware that requires no additional
large language model calls. Positioned between the retriever and the generator,
it performs stance detection, fixed balancing (2 supporting + 2 opposing + 1
neutral), and a debiasing prompt. Its online balancing rule contains no
thresholds or weights tuned on the test set.

Third, we provide a systematic evaluation pipeline covering four poisoning
concentrations, two poisoning directions, four generation models, and 210
controversial questions. The evaluation reports defense effects, efficiency
overhead, content-moderation blocks, refusals, and a comparison of evidence-selection mechanisms on shared retrieval results.

Fourth, we analyze how evidence selection shapes answer stance and examine boundary cases. The results
distinguish stance balance from content diversity and show how missing stance
categories and target-sensitive classification errors can affect the evidence
composition supplied to the generator.

\section{Related Work}\label{sec:related-work}

This section reviews three lines of research most directly related to
this study: RAG knowledge poisoning and its defenses, text stance
detection, and retrieval diversification and stance constraints.

\subsection{Retrieval-Augmented Generation and Its
Applications}\label{subsec:rag-applications}

Recent surveys characterize retrieval-augmented large language models as
comprising three main stages: retrieval, augmentation, and generation
\cite{ref01}. A system first retrieves query-relevant materials from external
knowledge sources, organizes them into a generation context, and then
uses a language model to answer the question based on that context.
External knowledge can be updated independently of model parameters and
can provide traceable evidence for answers, but it also makes generated
outputs substantially dependent on the content and quality of the
retrieved materials. For retrieval augmentation, AT-RAFT \cite{ref04} first
performs retrieval with a dual encoder and then reranks the results with
a cross-encoder; IIKM \cite{ref05} combines BM25 with vector retrieval and
applies a second round of ranking to the candidate results. Such
two-stage or hybrid retrieval can improve relevance and recall, but its
objective remains to identify ``more relevant content'' and does not
constrain the stance composition of the retrieval results. In other
words, even if every document is highly relevant to the query, the final
context may still cover only one side of a controversial issue.

RAG has been applied to specialized domains including aviation safety,
industrial logs, financial information extraction, educational question
answering, and medical assistance \cite{ref06,ref07,ref08,ref09,ref10}. RACI identifies causal
factors from aviation accident reports and improves retrieval robustness
through noise filtering and adversarial training; LogRAG retrieves
historical normal logs to support the review of anomaly detection
results; ESGReveal extracts structured financial information from
corporate reports; PLRTQA integrates answers dispersed across textbooks
through cross-document retrieval; and LLMind Chat restricts the
knowledge source of a medical assistance system to the authoritative
ICD-11 manual. Although these applications address different tasks, they
collectively demonstrate that generation quality depends on the
retrieval context and the reliability of the knowledge sources. In high-stakes domains such as aviation, finance, and
medicine, knowledge-base contamination may further affect downstream
judgments. Natural noise generally arises from imperfect retrieval,
whereas knowledge poisoning involves content actively constructed and
injected by an attacker so that it can be retrieved.

\subsection{RAG Knowledge Poisoning and
Defense}\label{subsec:rag-poisoning-defense}

\subsubsection{Knowledge Poisoning
Attacks}\label{subsubsec:knowledge-poisoning-attacks}

Knowledge poisoning allows an attacker to influence specific queries
solely by manipulating the knowledge base, without modifying the LLM.
PoisonedRAG \cite{ref11} systematically formulates RAG knowledge poisoning
as an optimization problem and proposes both black-box and white-box
variants. Its results showed that injecting five malicious texts per
target query into a knowledge base containing millions of documents
could achieve attack success rates of 90\%--97\%. This finding
demonstrates that a small number of injections can affect the retrieval
and generation pipeline without requiring the attacker to access model
parameters. BadRAG \cite{ref12} further constructs trigger-activated
adversarial passages through contrastive optimization to implement
retrieval backdoor attacks. Among its attack modes,
Selective-Fact-as-an-Attack (SFaaA), which guides the sentiment or
stance of outputs by selectively presenting facts, is the form most
closely related to the attack studied here.

This study differs from BadRAG/SFaaA in its research target. BadRAG focuses on retrieval backdoors and their trigger
mechanisms; SFaaA is one of its attack modes and does not separately
characterize and quantify systematic manipulation of the stance
distribution of retrieved materials. In contrast, this study
investigates stance poisoning that requires neither factual errors nor
special triggers and changes the stance of an answer solely through
one-sided but truthful arguments. Web crawling, user uploads, and
automated data pipelines in practical systems may all provide injection
channels, giving the assumption that an attacker can manipulate only the
knowledge base a realistic basis.

Topic-FlipRAG studies topic-level opinion manipulation across multiple related
queries and constructs retrieval-promoted adversarial documents through semantic
editing and adversarial triggers \cite{topicfliprag}. Our work differs in both
objective and mechanism. Rather than designing a stronger attack, we study an
online defense for factually grounded but one-sided document injection that does
not rely on specialized triggers. The proposed middleware uses a locally
executed stance classifier, a fixed evidence quota, and a debiasing instruction
to constrain the stance composition presented to the generator.

Most existing poisoning studies target answers whose correctness
can be determined and measure effects using answer accuracy or attack
success rate. Fact-checking, comparison with external knowledge, and
multi-document consistency checking accordingly rely on the premise that
the attack content contains verifiable errors. This premise does not
hold for materials that are factually correct but one-sided in stance:
each individual claim may be true, yet selective coverage can alter the
stance composition of the retrieval pool. Meanwhile, language models may
have their own tendencies on controversial issues \cite{ref13,ref14}. If the
model prior aligns with the direction of the retrieved materials, the
observed answer stance may reflect both influences.

\subsubsection{Existing Defense Methods and Their
Limitations}\label{subsubsec:existing-defense-limitations}

RobustRAG \cite{ref15} adopts an ``isolate-then-aggregate'' strategy. It
first prompts the LLM to generate an answer independently for each
retrieved document and then merges the results through secure
aggregation, which can provide robustness guarantees. This method avoids
interactions caused by directly mixing documents, but the number of
generation calls grows with the number of retrieved documents, resulting
in a high cost for real-time deployment. TrustRAG \cite{ref16} employs a
two-stage defense: it first uses K-means clustering to identify
documents that deviate from the main semantic groups and then combines
cosine similarity, ROUGE, and self-evaluation to handle suspicious
content and conflicts between internal and external knowledge. Its
decisions depend on semantic grouping and involve clustering parameters.
GRADA \cite{ref17}, by contrast, constructs an inter-document semantic
similarity graph and propagates scores in a PageRank-like manner to
strengthen densely connected document groups and suppress isolated
nodes, thereby reducing the influence of poisoned content through
reranking.

Stance-poisoned documents can be factually correct and topically consistent
with clean materials, so semantic outlier detection or graph-based filtering
may miss their collectively one-sided stance composition. Semantic grouping
is not equivalent to stance grouping. RobustRAG avoids this assumption but
requires per-document generation and aggregation, leaving a need for low-cost
defenses that directly constrain stance composition.

Another line of work adds behavioral constraints to prompts. Ganguli et
al. \cite{ref18} found that sufficiently large models trained with human
feedback possess a degree of moral self-correction and that
natural-language instructions can reduce certain stereotypes in model
outputs. Si et al. \cite{ref19} systematically investigated the effects of
prompting on model reliability and likewise observed improvements,
including reductions in social bias. Prompt-based methods are
inexpensive, decoupled from the model, and leave the retrieval process
unchanged, but they act only on the existing context: when the evidence is one-sided, a prompt cannot add materials for the missing side.

\subsection{Stance Detection, Reranking, and
Evaluation}\label{subsec:stance-detection-reranking-evaluation}

\subsubsection{Text Stance
Detection}\label{subsubsec:text-stance-detection}

Stance detection determines whether a text supports, opposes, or remains
neutral toward a specific topic or claim. Existing datasets vary
substantially in topic scope, label systems, and scale, and
generalization across domains and to topics unseen during training
remains an important challenge \cite{ref20}. Cross-topic generalization matters for an open knowledge base: the issues addressed by candidate
documents may not appear in the training data, so a detector must both
distinguish stances and handle unseen topics.

One approach to open-domain stance detection treats the text as a
premise and constructs hypotheses from the target topic and candidate
stances, thereby obtaining indirect supervision from the entailment,
contradiction, and neutral relations used in natural language inference
(NLI). Xu et al. \cite{ref21} map support, opposition, and neutrality to
these three relations, respectively, initializing RoBERTa-large with
MNLI before further fine-tuning it on stance data. Their work also notes
that stance detection and NLI are not fully equivalent and that direct
transfer remains difficult for open-domain and open-form topics.
Accordingly, this study fine-tunes the MNLI-initialized model on stance data (Section~\ref{subsec:stance-detection}).

\subsubsection{Diversity Reranking}\label{subsubsec:diversity-reranking}

A recent survey of result diversification classifies MMR as a greedy
post-processing method: the system incrementally constructs a result
list, selecting at each step a candidate that balances query relevance
against its difference from the already selected set \cite{ref22}. This
study implements the following scoring function:

\[d^*=\arg\max_{d\in R\setminus S}\left[\lambda\,Sim(d,q)-(1-\lambda)\max_{d_j\in S}Sim(d,d_j)\right].\]

Here, \(R\) denotes the candidate set, \(S\) the selected set, and
\(\lambda\) controls the trade-off between query relevance and content
dissimilarity. Related methods generally balance relevance, content dissimilarity, and aspect
coverage, but they do not directly constrain the stance composition of
the results.

Content diversity and stance balance are distinct constraint objectives.
Documents on the same side may still use different arguments and wording and thus remain dissimilar in the content space. Reducing content redundancy therefore does not directly limit the numbers of supporting, opposing, and neutral documents; this requires an explicit constraint on stance structure. Section~\ref{sec:baseline-comparison} compares the two constraints, and Section~\ref{sec:content-vs-stance} discusses the results.

\subsubsection{RAG Evaluation}\label{subsubsec:rag-evaluation}

RAG systems are typically evaluated in terms of both retrieval quality
and generation quality. RAGAS \cite{ref23} uses faithfulness to measure the
factual consistency of an answer with its context, answer relevance to
measure the match between an answer and its query, and context precision
and recall to assess the retrieval results and their coverage. These
metrics are suitable for determining whether an answer is evidence-based
and relevant and whether retrieval is effective, but they do not
directly measure stance skew in either the retrieved materials or the
generated answer. An answer may faithfully restate a one-sided context
while still performing well according to RAGAS metrics.

Poisoning studies commonly use attack success rate to measure whether a
target incorrect answer is induced, but this metric presupposes a
clearly defined erroneous target and cannot replace stance measurement
for controversial issues. Stance skew is an aggregate statistic over a
question set, and a single point estimate cannot represent sampling
uncertainty. Relevant NLP evaluation research recommends reporting
differences in method performance together with their bootstrap
confidence intervals \cite{ref24}, while statistical learning research also
treats the bootstrap as a standard resampling method \cite{ref25}. This
study therefore performs sampling with replacement at the topic level
and reports 95\% confidence intervals for continuous stance metrics and
differences between methods; the number of resamples and the interval calculation procedure are provided in Appendix~\ref{sec:paired-holm}.

\subsection{Research Gaps}\label{subsec:research-gaps}

Existing research leaves three gaps directly relevant to this study.
First, knowledge poisoning research has primarily focused on factual
errors and answers whose correctness can be determined. SFaaA addresses shifts
in tendency caused by selectively presented facts \cite{ref12}, and Topic-FlipRAG
demonstrates topic-level opinion manipulation across related queries
\cite{topicfliprag}. The remaining question addressed here is how to defend
online against factually grounded but one-sided injection without specialized triggers, in both poisoning directions. Second, methods such as TrustRAG and GRADA use
semantic anomalies or graph structures \cite{ref16,ref17}, whereas existing
result-diversification methods mainly constrain content dissimilarity or
the coverage of query aspects \cite{ref22}. None of these constraints
guarantees balance among supporting, opposing, and neutral documents in
the retrieval results. Third, the per-document generation used by
RobustRAG incurs a call cost that grows with the number of documents
\cite{ref15}, whereas low-cost prompting \cite{ref18,ref19} cannot alter the
composition of the evidence. Existing approaches therefore leave a gap
between deployment cost and structural constraints. On this basis, the
present study addresses stance measurement, explicit stance constraints
on retrieval results, and low-cost deployment.

\section{Method}\label{sec:method}

\subsection{Problem Formulation}\label{subsec:problem-formulation}

Given a controversial question \(q\) for which no single answer is
universally accepted and multiple positions are supported by evidence, a
retrieval-augmented generation (RAG) system first uses a retriever
\(\mathcal{R}\) to retrieve relevant documents
\(\mathcal{R}(q,\mathcal{D})\) from a document corpus \(\mathcal{D}\),
and then supplies a subset of these documents together with \(q\) to a
generator \(\mathcal{G}\) to produce an answer \(a\). For such
questions, answer quality cannot be assessed solely in terms of factual
correctness, and the stance of the answer may also be externally
manipulated.

Accordingly, we define stance poisoning as follows: an attacker injects
factually correct but highly one-sided documents into \(\mathcal{D}\) to
alter the distribution of evidence observed by the retriever and thereby
skew the generated answers. Because the attack targets evidence
selection rather than factuality, defenses predicated on determining
whether content is true or false are not applicable.

Our primary metric measures answer stance; Section~\ref{sec:clean-quality}
separately evaluates answer quality under clean conditions. For each answer \(a\), a separate
evaluation model (see Section~\ref{sec:experimental-setup} for its identity and for the isolation
measures ensuring that the evaluation model and the evaluated generation
model come from different model families) classifies it as pro, neutral,
or con, mapped to \(+1\), \(0\), and \(-1\), respectively. Averaging
over the evaluation question set \(\mathcal{Q}\) yields the stance score

\begin{equation}
\label{eq:stance-score}
s \;=\; \frac{1}{|\mathcal{Q}'|}\sum_{q \in \mathcal{Q}'} v(a_q), \qquad v(\cdot) \in \{+1,\,0,\,-1\}, \qquad s \in [-1,\,+1].
\end{equation}

where \(\mathcal{Q}' \subseteq \mathcal{Q}\) is the set of valid
questions, as defined below. A score of \(+1\) indicates that all
answers are supportive, whereas \(-1\) indicates the opposite; a score
close to \(0\) indicates that answers on the two sides are approximately
balanced in aggregate. The stance score characterizes the overall
tendency of a set of answers rather than the correctness of any
individual answer. In isolation, a single pro answer is not problematic;
the problem arises when a system systematically presents only one side
across an entire set of controversial questions.

Three additional outcome types require separate treatment:
\texttt{refuse} (the generator declines to take a position),
\texttt{filtered} (the answer is blocked by a content moderation
mechanism), and \texttt{unparsed} (the judging model\textquotesingle s
output cannot be parsed). None of these outcomes expresses a stance.
Encoding them as \(0\) would mechanically pull \(s\) toward neutrality
and conflate a change in stance with a decline in the rate of
stance-bearing responses. We therefore exclude all three from
\(\mathcal{Q}'\) and report their counts separately.

We evaluate three quantities. First, the
setting-A stance score and its 95\% confidence interval characterize
attack-induced skew. Second, the primary defense effect is the reduction in
absolute distance from 0 from no defense to the complete method,
$|s_A|-|s_C|$; a positive value means that the score moved closer to 0, and a
paired interval entirely above 0 indicates a clearly detected reduction in
deviation magnitude. Third, the setting-C stance score and its interval describe
residual skew after defense. A 95\% confidence interval represents the
uncertainty in the stance score under the current sample and resampling
procedure: an interval that excludes 0 supports a directionally clear aggregate
tendency, whereas an interval that includes 0 means that no directionally clear
residual is detected. Inclusion of 0 does not establish that the true bias is
exactly 0 or that outputs have achieved balance.

For A-to-C comparisons, only questions valid in both settings enter the strictly
paired analysis. Topic-cluster paired bootstrap intervals report effect sizes
and their uncertainty, while two-sided paired randomization tests provide
p-values. The eight A-to-C comparisons formed by two poisoning directions and
four concentrations constitute one test family and are adjusted using the Holm
procedure. This multiplicity adjustment changes the evidential assessment across
comparisons, not the direction of the effect-size estimates.

\subsection{Overall Architecture}\label{subsec:overall-architecture}

\begin{figure}[t]
\centering
\includegraphics[width=\linewidth]{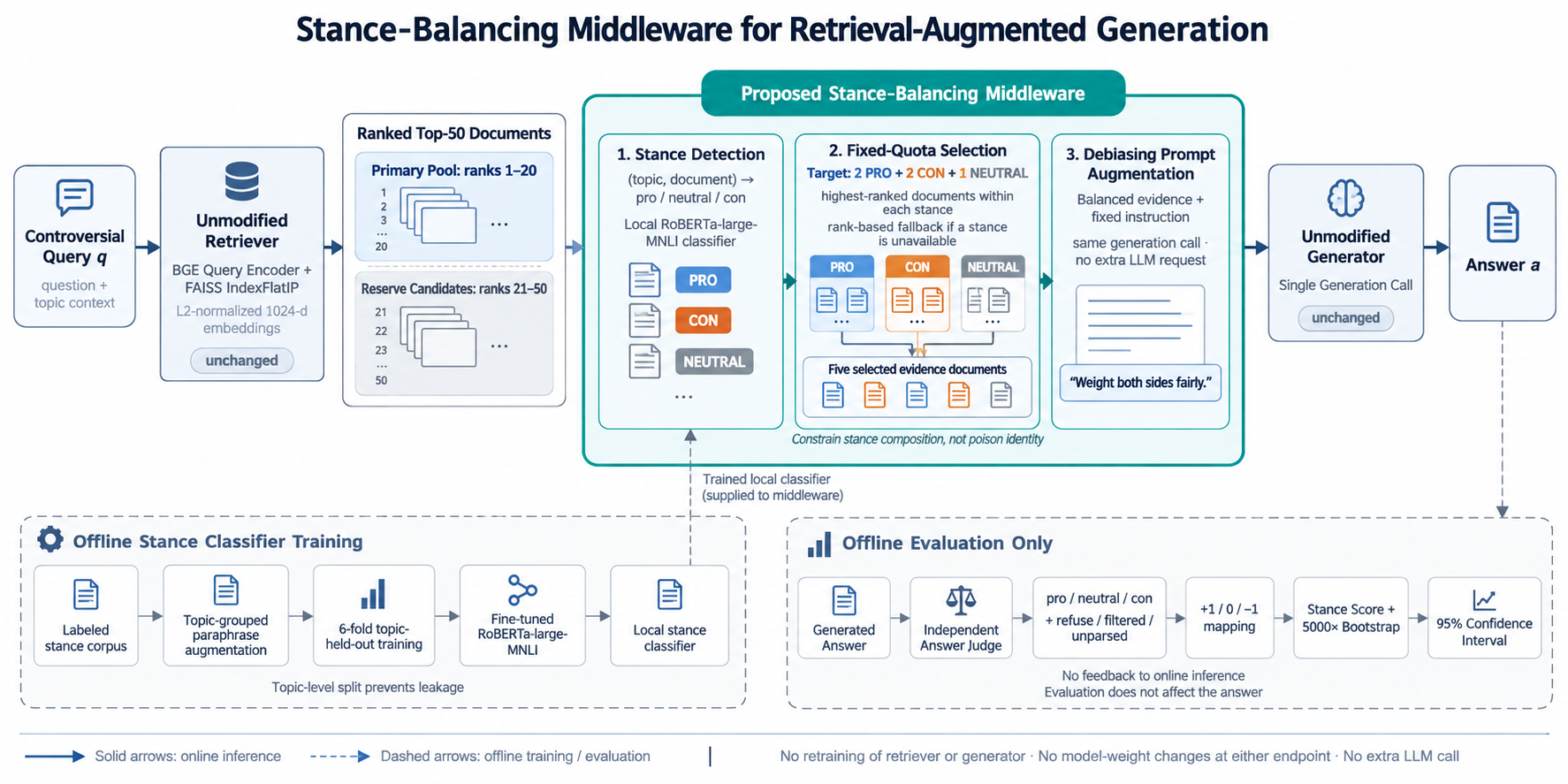}
\caption{Overall architecture: the Stance-Balancing Middleware is inserted between the retriever and the generator, with neither endpoint modified.}
\label{fig:architecture}
\end{figure}

We propose the Stance-Balancing Middleware, which is inserted between
the retriever and the generator. It receives the candidate document list
returned by the retriever, sequentially applies three
components---stance detection, fixed balancing, and a debiasing
prompt---and then passes a small, reselected set of documents together
with an additional instruction to the generator. The three components
form a unidirectional pipeline with no loops and require no multi-turn
interaction. The overall workflow is shown in Fig.~\ref{fig:architecture}.

The middleware reselects retrieved documents and adds an instruction without
modifying the retriever, generator, or document corpus. It preserves the existing
interfaces and requires no retraining of either endpoint.
Section~\ref{sec:cross-generator-generalization} evaluates compatibility with
four generators from different vendors.

\subsection{Component 1: Stance
Detection}\label{subsec:stance-detection}

The first component classifies each retrieved candidate document as supportive
(pro), opposing (con), or neutral with respect to the current question. It
identifies stance rather than factual correctness.

We use a locally executed, task-specific classifier to label candidate documents
without additional LLM calls. Section~\ref{sec:efficiency-analysis} measures its
latency relative to generation time. This document classifier is distinct from
\texttt{gpt-5-mini} in Section~\ref{sec:experimental-setup}, which labels generated
answers only during offline evaluation.

The classifier is fine-tuned from \texttt{FacebookAI/roberta-large-mnli}
(355 million parameters). This backbone has been trained on natural
language inference (NLI), whose task formulation is similar to stance
classification, making it a suitable initialization for our few-shot
fine-tuning setting. Classifier inference is performed entirely locally,
without calling any external API.

Training and evaluation use 6-fold topic-held-out cross-validation. Topics are
divided into six groups, and each classifier is trained on five groups and
evaluated on the held-out group. Thus, every document is classified by a model
that has not seen its topic during training, measuring cross-topic generalization
rather than topic-specific memorization.

The evaluation corpus contains 1,898 original documents (738 pro, 740 con, and
420 neutral). Each original document is paired with one encyclopedia-style
paraphrase for augmentation, with each pair kept in the same topic fold. The
classifier is trained for five epochs using AdamW with a learning rate of
\texttt{1e-5}, batch size 16, maximum length 256, and seed \texttt{20260831}; each
of the six folds holds out seven complete topics.

Under this protocol, the classifier achieves a three-class macro-F1 of 0.889 and
a target-stance classification accuracy of 0.905 on 588 injected documents. On
the pooled out-of-fold predictions, the Neutral class achieved an F1 of 0.973,
compared with 0.856 for pro and 0.848 for con; most errors were direct pro--con
confusions.

\subsection{Component 2: Fixed Stance Quota
(2+2+1)}\label{subsec:fixed-stance-quota}

The second component is the core of the Stance-Balancing Middleware and
determines the final set of documents supplied to the generator.

\textbf{Retrieval constants.} For each question, the retriever returns
the top \(K_{\text{search}} = 50\) ranked documents. The top
\(K_{\text{pool}} = 20\) constitute the primary candidate pool.
Documents ranked 21 through 50 are used only as supplementary sources
when documents of a particular stance are insufficient. The number of
documents ultimately supplied to the generator is fixed at \(F = 5\).

\textbf{Balancing rule.} After stance classification, the rule does not
directly pass the top 5 ranked documents to the generator. Instead, it
reselects 5 documents according to fixed quotas:

\begin{quote}
Step 1. Within the candidate pool (the top 20 documents), select by
retrieval rank, from highest to lowest, up to 2 pro documents, 2 con
documents, and 1 neutral document. Step 2. If any class does not meet
its quota within the candidate pool, continue down the retrieval ranking
to rank 50 and select additional documents from that class until its
quota is filled. Step 3. If the fallback search has exhausted all three
classes and fewer than 5 documents have been selected, disregard the
quotas and fill the remaining positions by retrieval rank, from highest
to lowest, until exactly five documents have been selected.
\end{quote}

When all three classes are sufficiently represented, this rule limits
the number of slots occupied by a single stance and does not depend on
identifying whether a document was injected by an attacker.

The 2+2+1 quota was not searched on the test set, and the online balancing rule
contains no thresholds or weights tuned on that set.

In the Step 3 case, the final set does not meet the 2+2+1 quota. Appendix~\ref{sec:failure-missing-neutral} provides examples and the distribution of this fallback case.

\subsection{Component 3: Debiasing
Prompt}\label{subsec:debiasing-prompt}

The third component adds a lightweight intervention on top of balancing.
An instruction is appended to the prompt supplied to the generator,
informing the model that the provided materials may be imbalanced in
stance. The model is instructed not to accept a
document\textquotesingle s conclusion merely because the document is
longer or uses stronger language, but instead to weigh the strongest
evidence from both sides equally. If the question is genuinely
contested, the answer should reflect that disagreement rather than adopt
the framing of either side.

All settings use a common base generation template that asks the model to answer
directly in 3--5 English sentences using only the retrieved documents, ground
each substantive claim in those documents, and neither refuse nor request more
information merely because the evidence is limited. Settings C and D insert the
same fixed debiasing instruction after the question and before the output
requirements, whereas settings A and B do not. The additional instruction neither
changes the selected documents nor adds a model call.
Appendix~\ref{sec:generation-prompts} provides the verbatim templates.

\subsection{Complexity and
Overhead}\label{subsec:complexity-overhead}

The three components have the following overhead profile. Stance
detection incurs linear cost in the number of candidate documents: one
small-model forward pass is performed for each candidate document. The
number of documents is determined by retrieval depth rather than corpus
size, and this is the only component with substantive computational
cost. Fixed balancing consists solely of ranking and counting logic,
involves no model inference, and has negligible runtime relative to
stance detection. The debiasing prompt is a constant-length text segment
that adds only a small number of tokens to the prompt and does not
increase the number of calls.

The method therefore has an important end-to-end overhead property: it
introduces no additional large-language-model calls. All computation in
the Stance-Balancing Middleware is performed by a local small model and
ranking logic. Consequently, it neither increases the number of calls to
the remote generation service nor incurs the corresponding cost
increase. This contrasts with defenses based on multi-round
verification, post-retrieval review, or multi-model voting, which
typically trade multiplicative increases in large-language-model calls
for robustness.

Measured CPU and GPU latency, together with their proportions relative
to the time required to generate one answer, are reported in Section~\ref{sec:efficiency-analysis}.

\section{Experiments}\label{sec:experiments}

\subsection{Experimental Setup}\label{sec:experimental-setup}

\subsubsection{Datasets}\label{sec:datasets}

The evaluation set contains 210 controversial questions spanning 42
topics, with five questions per topic. The topics were drawn from a
public list of controversial issues and cover areas including public
policy, technology ethics, and social institutions. Their common
characteristic is the absence of a single widely accepted answer.

The clean knowledge base consists of a subset of the Natural Questions corpus
containing 51,898 documents. Two direction-specific poisoned corpora were
constructed on this basis: all documents injected into the \texttt{pro} corpus
support the corresponding stance, whereas all documents injected into the
\texttt{con} corpus oppose it. The complete bidirectional set contains 588
injected documents, with 294 documents per direction. The four single-direction
knowledge bases contain 84, 168, 210, and 294 injected documents, respectively.
The separately constructed set of 420 neutral Wikipedia documents is not part of
the injected-document collection. Appendix~\ref{sec:measured-properties} provides
a descriptive comparison of word count and within-group nearest-neighbor BGE
cosine similarity. For concision, the tables below refer to the pro-poisoned and
con-poisoned corpora as the pro and con corpora, respectively.

Two human annotators independently assessed 252 of the 588 injected documents,
sampling three from each of the 84 topic--direction groups. Documents initially
flagged for factual issues or insufficient evidence were reviewed, and
disagreements were adjudicated. All 252 sampled documents were ultimately judged
to contain verifiable claims supported by the supplied sources and a clearly
one-sided stance matching the intended injection label.
Appendix~\ref{sec:human-document-validation} reports the criteria and results.

For topic $t$, let $n \in \{2,4,5,7\}$ be the number of documents injected in one
direction and let $N_t$ be the original number of ProCon documents. The realized
poisoning share is
\[
r_t=\frac{n}{N_t+n}, \qquad R=\frac{1}{42}\sum_{t=1}^{42}r_t.
\]
Thus, the share for each topic uses the number of injected documents in that
direction divided by the post-injection number of ProCon documents, and $R$ is
its macro-average over the 42 topics. The four target concentrations of 5\%,
10\%, 15\%, and 20\% yield realized macro-average shares of 6.4\%, 11.9\%, 14.4\%,
and 18.9\%, respectively. The gaps between target and realized shares arose from relevant documents already present in the clean corpus. The share covers only the ProCon documents of each topic; it excludes the rest of the knowledge base, including the neutral Wikipedia documents.
The main text reports the actual values throughout. All four levels are included
in the concentration-sweep attack and defense analyses; setting D is evaluated
only at the two highest levels in the ablation study.

\subsubsection{Baselines}\label{sec:baselines}

We consider two evidence-selection baselines, each selecting five documents from
the top-20 candidate pool. The proposed method also starts from this pool but can
extend its search to rank 50 when a stance quota is unmet
(Section~\ref{subsec:fixed-stance-quota}):

\begin{itemize}
\tightlist
\item
  \textbf{Random-5:} Five documents are sampled without replacement from
  the candidate pool using a fixed random seed of \texttt{20260909}.
\item
  \textbf{MMR (maximal marginal relevance):} Documents are selected
  greedily. At each step, MMR considers both relevance to the question
  and a penalty for excessive similarity to already selected documents,
  using the default trade-off coefficient $\lambda$ = 0.5. It represents a
  content-diversification approach in which the five selected documents
  are mutually nonredundant. MMR removes redundancy only at the content
  level and does not consider stance, which is precisely the distinction
  examined in this study. Document vectors are reconstructed and reused
  directly from the FAISS index. Because the index is an L2-normalized
  \texttt{IndexFlatIP}, inner products equal cosine similarity; the
  procedure introduces neither additional encoding nor additional model
  calls.
\end{itemize}

\subsubsection{Evaluation Metrics}\label{sec:evaluation-metrics}

The stance score is defined in Section~\ref{subsec:problem-formulation}.
Setting A characterizes attack-induced deviation, the strictly paired reduction
$|s_A|-|s_C|$ is the primary effect size for the complete method, and setting C
characterizes residual deviation after defense. Because questions within a topic
may be correlated, we resample whole topics when estimating confidence
intervals. Paired comparisons use questions valid in both settings.
Appendix~\ref{sec:paired-holm} provides the resampling and testing details.

\subsubsection{Implementation Details}\label{sec:implementation-details}

The retriever is \texttt{BAAI/bge-large-en-v1.5}~\cite{ref02}, which outputs
1,024-dimensional vectors. After L2 normalization, the vectors are
indexed using FAISS \texttt{IndexFlatIP}~\cite{ref03}, so the inner product
equals cosine similarity. The retrieval depth, candidate-pool size, and
number of documents ultimately supplied to the generator are 50, 20, and
5, respectively, consistent with
Section~\ref{subsec:fixed-stance-quota}. The main concentration-sweep,
ablation, and selection-baseline experiments use \texttt{claude-opus-4-8} as the
generator; Section~\ref{sec:cross-generator-generalization} separately replaces
it with the four generators listed there. Sampling parameters not explicitly
supplied by the client follow the gateway defaults.

All experiments in this section use the four settings defined in
Table~\ref{tab:experimental-settings}.

\begin{longtable}[]{@{}>{\raggedright\arraybackslash}p{0.08\linewidth}>{\raggedright\arraybackslash}p{0.18\linewidth}>{\raggedright\arraybackslash}p{0.42\linewidth}>{\raggedright\arraybackslash}p{0.24\linewidth}@{}}
\caption{Definitions of the four experimental settings}\label{tab:experimental-settings}\\
\toprule\noalign{}
Setting & Name & Selection of the five documents supplied to the
generator & Prompt \\
\midrule\noalign{}
\endfirsthead
\toprule
Setting & Name & Selection of the five documents supplied to the
generator & Prompt \\
\midrule
\endhead
\bottomrule\noalign{}
\endlastfoot
\textbf{A} & No defense & Directly supply the top five retrieved
documents & No additional instruction \\
\textbf{B} & Balancing (no prompt) & 2 pro + 2 con + 1 neutral & No additional
instruction \\
\textbf{C} & Complete method (balancing + prompt) & 2 pro + 2 con + 1 neutral & Append a
debiasing instruction \\
\textbf{D} & Prompt only & Top five retrieved documents, as in A &
Append the same debiasing instruction as in C \\
\end{longtable}

Settings B and C both use stance-detection results to perform fixed balancing;
C additionally applies the debiasing prompt. Setting A characterizes the
attacked state without defense, setting C is the complete Stance-Balancing
Middleware advocated in this study, and settings B and D isolate the
contribution of each component.

The answer-stance judge is \texttt{gpt-5-mini} in all experiments; its fallback chain, which excludes the current generator, was never triggered. Judging is
used only for offline evaluation, and its cost is excluded from the online
latency reported in Section~\ref{sec:efficiency-analysis}. The answer-stance judge
was additionally checked against human-reviewed reference labels;
Appendix~\ref{sec:evaluation-human-checks} reports the validation basis.

\subsection{Main Results}\label{sec:main-results}

\subsubsection{Attack Effectiveness}\label{sec:attack-effectiveness}

\begin{figure*}[t]
\centering
\includegraphics[width=\textwidth]{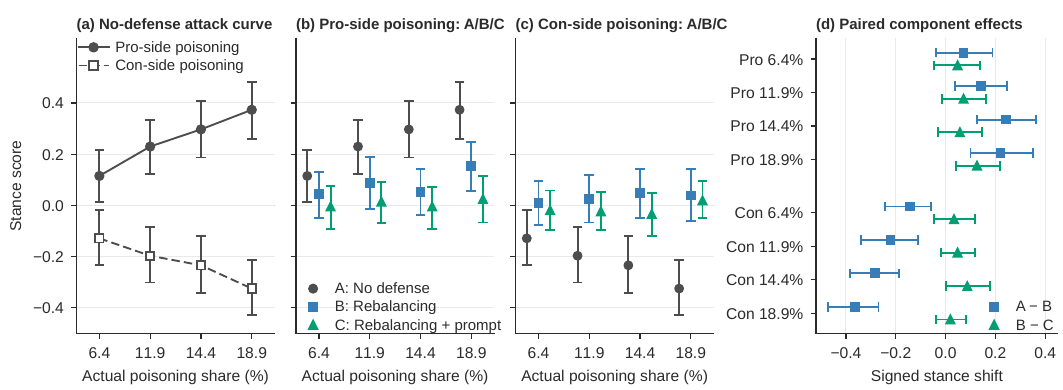}
\caption{Attack effectiveness across poisoning concentrations and defense effectiveness under pro- and con-side poisoning.}
\label{fig:attack-defense}
\end{figure*}

This section examines whether stance poisoning itself constitutes an
effective attack. Figure~\ref{fig:attack-defense}(a) shows how the attack varies with
concentration, and Table~\ref{tab:attack-effectiveness} reports stance scores for setting A (no
defense) under four poisoning concentrations and two poisoning
directions.

\begin{longtable}[]{@{}>{\raggedright\arraybackslash}p{0.19\linewidth}>{\raggedright\arraybackslash}p{0.17\linewidth}>{\raggedright\arraybackslash}p{0.28\linewidth}>{\raggedright\arraybackslash}p{0.28\linewidth}@{}}
\caption{Attack effectiveness: stance scores and 95\% confidence intervals without defense (210 questions per cell)}\label{tab:attack-effectiveness}\\
\toprule\noalign{}
Target concentration & Poisoning share & Pro-poisoned corpus &
Con-poisoned corpus \\
\midrule\noalign{}
\endfirsthead
\toprule
Target concentration & Poisoning share & Pro-poisoned corpus &
Con-poisoned corpus \\
\midrule
\endhead
\bottomrule\noalign{}
\endlastfoot
5\% & 6.4\% & +0.115 [+0.014, +0.215] & $-$0.129 [$-$0.234, $-$0.019] \\
10\% & 11.9\% & +0.230 [+0.124, +0.335] & $-$0.197 [$-$0.301, $-$0.086] \\
15\% & 14.4\% & +0.297 [+0.187, +0.409] & $-$0.234 [$-$0.343, $-$0.119] \\
\textbf{20\%} & \textbf{18.9\%} & \textbf{+0.373 [+0.260, +0.483]} &
\textbf{$-$0.325 [$-$0.429, $-$0.215]} \\
\end{longtable}

The stance-score magnitude increases monotonically with concentration in both
directions, and each shift aligns with the poisoning direction. All eight
confidence intervals exclude 0, providing additional evidence that the
attack-induced skew is directionally distinguishable even at the lowest actual
proportion of 6.4\%. Under this experimental setup, a mean topic-level poisoning
share below 20\% within the post-injection ProCon documents is sufficient to
shift answers overall toward the corresponding direction.

\subsubsection{Defense Effectiveness}\label{sec:defense-effectiveness}

Figure~\ref{fig:attack-defense}(b) and (c) show stance scores for settings A, B, and C, while
Figure~\ref{fig:attack-defense}(d) shows the signed stance shifts A$-$B and B$-$C. Table~\ref{tab:defense-effectiveness} reports
the corresponding numerical results for the same eight cells.

\begin{longtable}[]{@{}>{\raggedright\arraybackslash}p{0.08\linewidth}>{\raggedright\arraybackslash}p{0.14\linewidth}*{3}{>{\raggedright\arraybackslash}p{0.225\linewidth}}@{}}
\caption{Stance scores and 95\% confidence intervals for no defense, balancing without a prompt, and the complete method}\label{tab:defense-effectiveness}\\
\toprule\noalign{}
Corpus & Poisoning share & A: No defense & B: Balancing (no prompt) & C:
Complete method (balancing + prompt) \\
\midrule\noalign{}
\endfirsthead
\toprule
Corpus & Poisoning share & A: No defense & B: Balancing (no prompt) & C:
Complete method (balancing + prompt) \\
\midrule
\endhead
\bottomrule\noalign{}
\endlastfoot
Pro & 6.4\% & +0.115 [+0.014, +0.215] & +0.043 [$-$0.048, +0.129] & $-$0.005 [$-$0.091, +0.077] \\
Pro & 11.9\% & +0.230 [+0.124, +0.335] & +0.086 [$-$0.014, +0.190] & +0.014 [$-$0.068, +0.092] \\
Pro & 14.4\% & +0.297 [+0.187, +0.409] & +0.053 [$-$0.038, +0.144] & $-$0.005 [$-$0.091, +0.072] \\
Pro & \textbf{18.9\%} & +0.373 [+0.260, +0.483] & +0.153 [+0.057, +0.249] & \textbf{+0.024 [$-$0.067, +0.113]} \\
Con & 6.4\% & $-$0.129 [$-$0.234, $-$0.019] & +0.010 [$-$0.077, +0.097] & $-$0.019 [$-$0.096, +0.058] \\
Con & 11.9\% & $-$0.197 [$-$0.301, $-$0.086] & +0.024 [$-$0.067, +0.120] & $-$0.024 [$-$0.096, +0.052] \\
Con & 14.4\% & $-$0.234 [$-$0.343, $-$0.119] & +0.048 [$-$0.048, +0.144] & $-$0.034 [$-$0.119, +0.049] \\
Con & \textbf{18.9\%} & $-$0.325 [$-$0.429, $-$0.215] & +0.038 [$-$0.062, +0.144] & \textbf{+0.019 [$-$0.049, +0.095]} \\
\end{longtable}

Across all eight concentration--direction settings, the complete method reduced
absolute stance deviation by an estimated 0.106--0.350.
Table~\ref{tab:defense-effectiveness} shows the remaining stance scores, and its
continuation (Table~\ref{tab:defense-shifts}) reports the strictly paired A-to-C
effects. All eight setting-C intervals include 0, indicating no directionally
clear residual in this concentration sweep.

\begin{longtable}[]{@{}>{\raggedright\arraybackslash}p{0.08\linewidth}>{\raggedright\arraybackslash}p{0.15\linewidth}>{\raggedright\arraybackslash}p{0.335\linewidth}>{\raggedright\arraybackslash}p{0.335\linewidth}@{}}
\caption{Strictly paired effects of the complete method (A$\rightarrow$C) and 95\% confidence intervals (continuation of Table~\ref{tab:defense-effectiveness})}\label{tab:defense-shifts}\\
\toprule\noalign{}
Corpus & Poisoning share & Signed A$\rightarrow$C stance shift [95\% CI] &
Reduction in absolute distance from 0 [95\% CI] \\
\midrule\noalign{}
\endfirsthead
\toprule
Corpus & Poisoning share & Signed A$\rightarrow$C stance shift [95\% CI] &
Reduction in absolute distance from 0 [95\% CI] \\
\midrule
\endhead
\bottomrule\noalign{}
\endlastfoot
Pro & 6.4\% & +0.115 [+0.005, +0.226] & +0.106 [$-$0.038, +0.181] \\
Pro & 11.9\% & +0.216 [+0.110, +0.324] & +0.216 [+0.081, +0.300] \\
Pro & 14.4\% & +0.303 [+0.180, +0.433] & +0.293 [+0.139, +0.383] \\
Pro & 18.9\% & +0.350 [+0.232, +0.466] & +0.350 [+0.211, +0.447] \\
Con & 6.4\% & $-$0.111 [$-$0.228, +0.005] & +0.111 [$-$0.015, +0.205] \\
Con & 11.9\% & $-$0.173 [$-$0.273, $-$0.072] & +0.173 [+0.048, +0.262] \\
Con & 14.4\% & $-$0.203 [$-$0.300, $-$0.106] & +0.203 [+0.082, +0.295] \\
Con & 18.9\% & $-$0.351 [$-$0.442, $-$0.263] & +0.313 [+0.153, +0.411] \\
\end{longtable}

Note: The signed shift is $s_A-s_C$; its sign indicates movement along the stance
axis. The reduction in absolute distance is $|s_A|-|s_C|$, for which a positive
value indicates movement closer to 0. Each row uses only questions valid in both
A and C. Differences and intervals are computed from unrounded question-level
results and may therefore differ by 0.001 from direct subtraction of the
three-decimal displayed values.

The six moderate-to-high concentration comparisons remain supported after Holm
adjustment, while the two lowest-concentration comparisons have weaker adjusted
evidence. Assigning each refusal to con, neutral, or pro preserves the direction
of reduction in all eight settings, with intervals above 0 in the six
moderate-to-high settings. Together, these results support the complete
middleware's effectiveness in reducing stance deviation across both poisoning
directions and all tested concentrations, with stronger evidence at moderate and
high concentrations.

\subsection{Ablation Study}\label{sec:ablation-study}

To separate the role of prompting from that of balancing, we add setting D,
which retains the evidence selection used in setting A (directly supplying the
top five retrieved documents) and adds only the debiasing prompt. All four
settings were evaluated at the two highest concentrations. Point estimates are
reported in Table~\ref{tab:ablation}.

\begin{longtable}[]{@{}>{\raggedright\arraybackslash}p{0.08\linewidth}>{\raggedright\arraybackslash}p{0.15\linewidth}*{4}{>{\raggedright\arraybackslash}p{0.16\linewidth}}@{}}
\caption{Four-setting ablation comparing balancing alone, prompting alone, and their combination}\label{tab:ablation}\\
\toprule\noalign{}
Corpus & Poisoning share & A: No defense & B: Balancing (no prompt) & C:
Complete method (balancing + prompt) & \textbf{D: Prompt only} \\
\midrule\noalign{}
\endfirsthead
\toprule
Corpus & Poisoning share & A: No defense & B: Balancing (no prompt) & C:
Complete method (balancing + prompt) & \textbf{D: Prompt only} \\
\midrule
\endhead
\bottomrule\noalign{}
\endlastfoot
Pro & 14.4\% & +0.297 & +0.053 & $-$0.005 & \textbf{+0.167} \\
Pro & 18.9\% & +0.373 & +0.153 & +0.024 & \textbf{+0.256} \\
Con & 14.4\% & $-$0.234 & +0.048 & $-$0.034 & \textbf{$-$0.215} \\
Con & \textbf{18.9\%} & \textbf{$-$0.325} & \textbf{+0.038} &
\textbf{+0.019} & \textbf{$-$0.291} \\
\end{longtable}

Note: Values are point estimates; Table~\ref{tab:defense-effectiveness} reports the intervals for A, B, and C.

In point estimates, prompting alone reduces the absolute score by 0.019--0.130, balancing alone by 0.186--0.287, and the complete method leaves the smallest absolute residual (0.005--0.034) in all four settings. Balancing is therefore the main source of correction, and prompting adds a further reduction.

\subsection{Generalization across Generators}\label{sec:cross-generator-generalization}

\begin{figure*}[t]
\centering
\includegraphics[width=\textwidth]{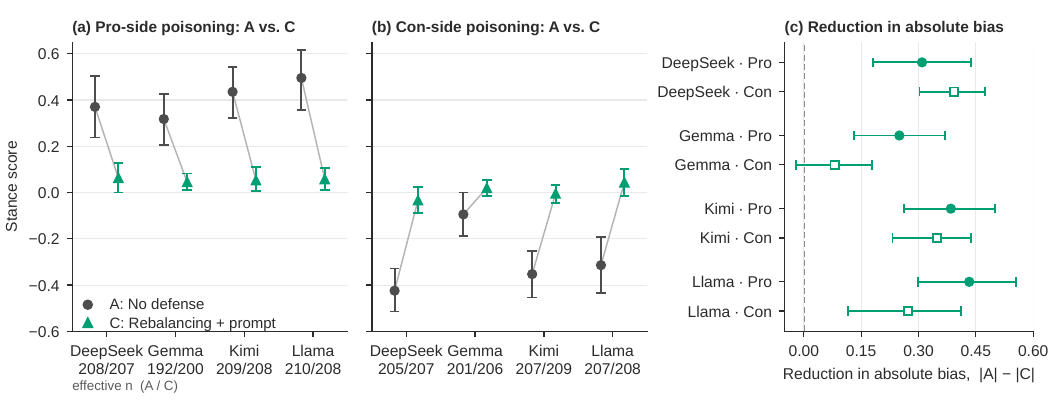}
\caption{Cross-generator stance scores before and after applying the complete defense at an actual poisoning proportion of 18.9\%.}
\label{fig:cross-model}
\end{figure*}

In this experiment, only the generator is changed; the questions, index,
retrieval results, stance detection, balancing code, and judge model are
held fixed. The experiment covers four providers---DeepSeek, Meta,
Moonshot, and Google---and compares A with C under both poisoning
directions at a poisoning share of 18.9\%. Generator fallback is disabled.
Cross-model results are presented in Figure~\ref{fig:cross-model} and Table~\ref{tab:cross-generator}.

\begin{longtable}[]{@{}>{\raggedright\arraybackslash}p{0.22\linewidth}*{4}{>{\raggedright\arraybackslash}p{0.17\linewidth}}@{}}
\caption{Cross-generator stance scores and 95\% confidence intervals (poisoning share: 18.9\%)}\label{tab:cross-generator}\\
\toprule\noalign{}
Generator & Pro A & Pro C & Con A & Con C \\
\midrule\noalign{}
\endfirsthead
\toprule
Generator & Pro A & Pro C & Con A & Con C \\
\midrule
\endhead
\bottomrule\noalign{}
\endlastfoot
DeepSeek-V4-Pro & +0.370 [+0.238, +0.505] & +0.063 [+0.000, +0.126] & $-$0.424 [$-$0.514, $-$0.328] & $-$0.034 [$-$0.088, +0.024] \\
Gemma-4-31B & +0.318 [+0.204, +0.426] & +0.045 [+0.010, +0.082] & $-$0.095 [$-$0.188, +0.000] & +0.019 [$-$0.014, +0.053] \\
Kimi-K2.6 & +0.435 [+0.321, +0.543] & +0.053 [+0.005, +0.110] & $-$0.353 [$-$0.454, $-$0.252] & $-$0.005 [$-$0.043, +0.033] \\
Llama-4-Maverick-17B & +0.495 [+0.357, +0.614] & +0.058 [+0.010, +0.106] & $-$0.314 [$-$0.435, $-$0.191] & +0.043 [$-$0.014, +0.100] \\
\end{longtable}

All four evaluated models exhibit setting-A shifts that align with the poisoning
direction. In all eight model--direction combinations, the signed A$-$C interval excludes 0 and the score moves along the axis in the direction that counteracts
the attack; 7/8 combinations also have an interval excluding 0 for the reduction
in absolute distance from 0. Under pro-side poisoning, the complete method reduces
deviation by more than 80\% from the setting-A values, with setting-C point
estimates from +0.045 to +0.063. Small positive residual point estimates remain;
their intervals exclude 0 in three models and include 0 in DeepSeek. Under
con-side poisoning, all four setting-C intervals include 0.
Section~\ref{sec:scope-limitations} discusses the remaining pro-side residuals.

Gemma-4-31B under con-side poisoning is the sole combination whose reduction in
absolute distance from 0 is not clearly detected: +0.081 [$-$0.021, +0.179],
despite a signed A$-$C shift of $-$0.121 [$-$0.207, $-$0.036]. Its setting-A score
is $-$0.095, compared with $-$0.314 to $-$0.424 for the other models. Counts of blocked
responses, refusals, and valid samples for each model are shown in
Table~\ref{tab:missing-responses}.

\begin{longtable}[]{@{}>{\raggedright\arraybackslash}p{0.21\linewidth}>{\raggedright\arraybackslash}p{0.08\linewidth}*{6}{>{\raggedright\arraybackslash}p{0.095\linewidth}}@{}}
\caption{Blocked-response and refusal statistics (denominator: 210 questions per cell)}\label{tab:missing-responses}\\
\toprule\noalign{}
Generator & Corpus & A blocked & A refused & A included & C blocked & C
refused & C included \\
\midrule\noalign{}
\endfirsthead
\toprule
Generator & Corpus & A blocked & A refused & A included & C blocked & C
refused & C included \\
\midrule
\endhead
\bottomrule\noalign{}
\endlastfoot
DeepSeek-V4-Pro & Pro & 1 & 1 & 208 & 1 & 2 & 207 \\
DeepSeek-V4-Pro & Con & 1 & 4 & 205 & 0 & 3 & 207 \\
Llama-4-Maverick-17B & Pro & 0 & 0 & 210 & 0 & 2 & 208 \\
Llama-4-Maverick-17B & Con & 0 & 3 & 207 & 0 & 2 & 208 \\
Kimi-K2.6 & Pro & 1 & 0 & 209 & 1 & 1 & 208 \\
Kimi-K2.6 & Con & 2 & 1 & 207 & 0 & 1 & 209 \\
\textbf{Gemma-4-31B} & Pro & \textbf{11} & \textbf{7} & \textbf{192} & 3
& 7 & 200 \\
Gemma-4-31B & Con & 1 & \textbf{8} & 201 & 1 & 3 & 206 \\
\end{longtable}

Gemma-4-31B has the most exclusions (up to 18 of 210 per cell); each other model excludes at most 5 per cell. Across the four tested
generators, the consistent corrective movement without model-specific adjustment
supports applicability beyond a single generator.

\subsection{Comparison of Selection Mechanisms: Fixed Balancing, Random Selection, and MMR}\label{sec:baseline-comparison}

This section compares the rules used to select the final five documents. Fixed
balancing, random selection, and MMR use the common base prompt without the
debiasing instruction, so the proposed method participates as setting B. The
methods share the 210 questions, index, retrieval results, initial top-20
candidate pool, generator, and judge model. Random-5 and MMR select within the
top 20, whereas the balancing rule permits extension to rank 50 when needed
(Section~\ref{subsec:fixed-stance-quota}). Question IDs are aligned across all
210 questions; paired score comparisons retain only questions valid in both
methods.

\begin{longtable}[]{@{}>{\raggedright\arraybackslash}p{0.07\linewidth}>{\raggedright\arraybackslash}p{0.20\linewidth}>{\raggedright\arraybackslash}p{0.25\linewidth}>{\raggedright\arraybackslash}p{0.10\linewidth}>{\raggedright\arraybackslash}p{0.29\linewidth}@{}}
\caption{Comparison of evidence-selection methods with shared retrieval results, generator, and judge (poisoning share: 18.9\%)}\label{tab:selection-baselines}\\
\toprule\noalign{}
Corpus & Method & Stance score {[}95\% CI{]} & Included & Paired
difference vs. B {[}95\% CI{]} \\
\midrule\noalign{}
\endfirsthead
\toprule
Corpus & Method & Stance score {[}95\% CI{]} & Included & Paired
difference vs. B {[}95\% CI{]} \\
\midrule
\endhead
\bottomrule\noalign{}
\endlastfoot
Pro & \textbf{B: Balancing (no prompt, ours)} & \textbf{+0.153 [+0.057, +0.249]} & 209 & --- \\
Pro & Random-5 & +0.330 [+0.219, +0.438] & 209 & +0.178 [+0.024, +0.321] * \\
Pro & MMR & +0.197 [+0.110, +0.285] & 208 & +0.043 [$-$0.062, +0.149] $\dagger$ \\
Con & \textbf{B: Balancing (no prompt, ours)} & \textbf{+0.038 [$-$0.062, +0.144]} & 209 & --- \\
Con & Random-5 & $-$0.234 [$-$0.332, $-$0.143] & 209 & $-$0.273 [$-$0.389, $-$0.168] * \\
Con & MMR & $-$0.148 [$-$0.263, $-$0.038] & 209 & $-$0.187 [$-$0.311, $-$0.071] * \\
\end{longtable}

Note: * denotes a paired-difference confidence interval that excludes 0;
$\dagger$ denotes an interval that includes 0. Neither baseline produced any content-moderation blocks.

Balancing without a prompt shows a clear advantage over random selection in both
directions; Random-5 remains at $-$0.234 in the con corpus. Relative to MMR, the
advantage is clear under con-side poisoning, with a paired difference of $-$0.187
[$-$0.311, $-$0.071]. Under pro-side poisoning, the paired difference from MMR is +0.043 [$-$0.062, +0.149].

Table~\ref{tab:poisoned-document-count} provides mechanism-level evidence by reporting the mean number
of poisoned documents among the five documents actually supplied to the
generator for each method, averaged over 210 questions.

\begin{longtable}[]{@{}>{\raggedright\arraybackslash}p{0.55\linewidth}>{\raggedright\arraybackslash}p{0.19\linewidth}>{\raggedright\arraybackslash}p{0.19\linewidth}@{}}
\caption{Mean number of poisoned documents among the five supplied per question}\label{tab:poisoned-document-count}\\
\toprule\noalign{}
Method & Pro corpus & Con corpus \\
\midrule\noalign{}
\endfirsthead
\toprule
Method & Pro corpus & Con corpus \\
\midrule
\endhead
\bottomrule\noalign{}
\endlastfoot
A: No defense (directly supply the top five) & 1.743 & 1.424 \\
\textbf{B / C: Proposed balancing} & \textbf{1.181} & \textbf{0.986} \\
Random-5 & 1.267 & 1.152 \\
\textbf{MMR} & \textbf{1.048} & \textbf{0.900} \\
\end{longtable}

MMR supplies fewer poisoned documents than the proposed method: 1.048
versus 1.181 for the pro corpus and 0.900 versus 0.986 for the con
corpus. Nevertheless, its con-side bias is greater ($-$0.148 versus
+0.038). This inverse ordering indicates that poison count alone does not
explain the final stance bias; Section~\ref{sec:content-vs-stance} discusses the
mechanism.

\subsection{Efficiency Analysis}\label{sec:efficiency-analysis}

This section quantifies the practical overhead of the middleware. All
latency measurements use the same 30 questions, index, batch size of 32,
and maximum sequence length of 256, with row-wise alignment; the ratios
are therefore comparable. None of the measurements consumes any API
calls.

Table~\ref{tab:detection-latency} reports detection latency as the number of candidate documents
varies.

\begin{longtable}[]{@{}>{\raggedright\arraybackslash}p{0.22\linewidth}*{3}{>{\raggedright\arraybackslash}p{0.13\linewidth}}>{\raggedright\arraybackslash}p{0.28\linewidth}@{}}
\caption{Stance-detection latency by number of candidate documents (median, $n = 30$)}\label{tab:detection-latency}\\
\toprule\noalign{}
Documents classified & CPU & GPU & GPU speedup & Share of generation
time (CPU / GPU) \\
\midrule\noalign{}
\endfirsthead
\toprule
Documents classified & CPU & GPU & GPU speedup & Share of generation
time (CPU / GPU) \\
\midrule
\endhead
\bottomrule\noalign{}
\endlastfoot
1 & 63.2 ms & 10.6 ms & 6.0$\times$ & 1.07\% / 0.18\% \\
5 & 237.1 ms & 22.8 ms & 10.4$\times$ & 4.02\% / 0.39\% \\
10 & 348.2 ms & 48.9 ms & 7.1$\times$ & 5.91\% / 0.83\% \\
\textbf{20 (actual deployment)} & \textbf{954.2 ms} & \textbf{88.8 ms} &
\textbf{10.7$\times$} & \textbf{16.2\% / 1.51\%} \\
32 & 2522.5 ms & 222.2 ms & 11.4$\times$ & 42.8\% / 3.77\% \\
50 (full recall set) & 3848.3 ms & 337.8 ms & 11.4$\times$ & 65.3\% / 5.73\% \\
\end{longtable}

Twenty documents represent the standard deployment setting. Fifty
documents comprise the full retrieval set and are all classified only
when a stance category is undersupplied and ranks 21--50 must be
accessed; this row therefore represents a worst-case upper bound rather
than routine overhead. The remaining rows show the latency curve as
candidate-set size varies.

The 88.8 ms GPU latency for 20 documents equals 1.51\% of the median remote-generation latency of \texttt{claude-opus-4-8} (5,894.3 ms; p10 4,783.6 ms; p90 7,381.7 ms) and 1.9\% of its p10 value. The generation latency comes from separate benchmark runs with matched inputs, so this ratio is a scale comparison rather than a per-query paired measurement.

Latencies are batch totals; batches of 32 are padded to their longest text, so
latency varies nonlinearly with document count.

Table~\ref{tab:routing-latency} reports an independent measurement. Its 20-document results
differ by approximately 1\% from the corresponding row in Table~\ref{tab:detection-latency} (CPU:
954.2 versus 945.3; GPU: 88.8 versus 89.7).

\begin{longtable}[]{@{}>{\raggedright\arraybackslash}p{0.31\linewidth}>{\raggedright\arraybackslash}p{0.12\linewidth}>{\raggedright\arraybackslash}p{0.12\linewidth}>{\raggedright\arraybackslash}p{0.25\linewidth}>{\raggedright\arraybackslash}p{0.12\linewidth}@{}}
\caption{$2\times2$ decomposition of six-fold routing versus a single model and 20 versus 50 documents (median, $n = 30$)}\label{tab:routing-latency}\\
\toprule\noalign{}
Configuration & CPU & GPU & Share of generation time (CPU $\rightarrow$ GPU) & GPU
speedup \\
\midrule\noalign{}
\endfirsthead
\toprule
Configuration & CPU & GPU & Share of generation time (CPU $\rightarrow$ GPU) & GPU
speedup \\
\midrule
\endhead
\bottomrule\noalign{}
\endlastfoot
\textbf{Single model / 20 documents (actual deployment)} & 945.3 ms &
\textbf{89.7 ms} & 16.0\% $\rightarrow$ \textbf{1.52\%} & \textbf{10.5$\times$} \\
Six-fold routing / 20 documents & 929.4 ms & 98.3 ms & 15.8\% $\rightarrow$ 1.67\% &
9.5$\times$ \\
Single model / 50 documents & 3900.1 ms & 337.4 ms & 66.2\% $\rightarrow$ 5.72\% &
11.6$\times$ \\
Six-fold routing / 50 documents (most conservative) & 3932.9 ms & 338.3
ms & 66.7\% $\rightarrow$ 5.74\% & 11.6$\times$ \\
\end{longtable}

Table~\ref{tab:routing-latency} shows GPU speedups of 9.5$\times$--11.6$\times$
across the tested configurations. Six-fold routing prevents topic leakage during
evaluation; deployment uses a single classifier. These measurements support integrating local stance detection into
the RAG pipeline with a small latency cost relative to generation.

\subsection{Robustness under Misclassification}\label{sec:misclassification-robustness}

Section~\ref{subsec:stance-detection} reports the stance detector\textquotesingle s
misclassification rate. This section analyzes how those errors affect
method effectiveness.

\begin{figure*}[t]
\centering
\includegraphics[width=\textwidth]{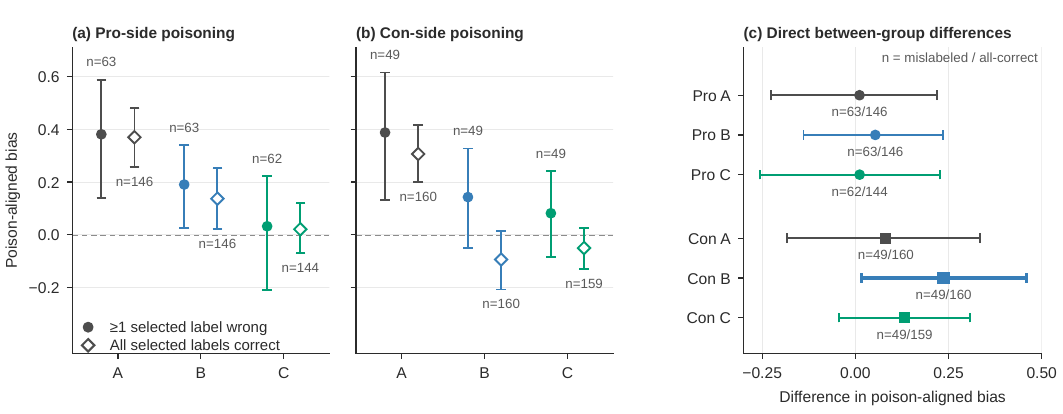}
\caption{Direct between-group differences for groups with and without detection errors.}
\label{fig:misclassification}
\end{figure*}

Grouping questions according to whether at least one stance among the
five ultimately supplied documents was misclassified yields 64/210
(30.5\%) for the pro corpus and 49/210 (23.3\%) for the con corpus. Figure~\ref{fig:misclassification} and Table~\ref{tab:misclassification} compare these groups. The value 210 is the number of
questions before grouping. Each setting still excludes blocked, refused,
and unparseable cases according to Section~\ref{subsec:problem-formulation}, so the effective \(n\)
values in Table~\ref{tab:misclassification} are slightly smaller and vary by setting.

\begin{longtable}[]{@{}>{\raggedright\arraybackslash}p{0.065\linewidth}>{\raggedright\arraybackslash}p{0.15\linewidth}>{\raggedright\arraybackslash}p{0.19\linewidth}>{\raggedleft\arraybackslash}p{0.085\linewidth}>{\raggedleft\arraybackslash}p{0.13\linewidth}>{\raggedright\arraybackslash}p{0.18\linewidth}>{\raggedright\arraybackslash}p{0.085\linewidth}@{}}
\caption{Poison-aligned bias and 95\% confidence intervals grouped by the presence of misclassification}\label{tab:misclassification}\\
\toprule\noalign{}
Corpus & Setting & Group & Ques-\newline tions & Poison-aligned bias & 95\%
confidence interval & Includes 0? \\
\midrule\noalign{}
\endfirsthead
\toprule
Corpus & Setting & Group & Ques-\newline tions & Poison-aligned bias & 95\%
confidence interval & Includes 0? \\
\midrule
\endhead
\bottomrule\noalign{}
\endlastfoot
Pro & A: No defense & With misclassification & 63 & +0.381 & [+0.140, +0.588] & No \\
Pro & A: No defense & All correct & 146 & +0.370 & [+0.258, +0.479] & No \\
Pro & B: Balancing (no prompt) & With misclassification & 63 & +0.190 & [+0.025, +0.339] & No \\
Pro & B: Balancing (no prompt) & All correct & 146 & +0.137 & [+0.022, +0.252] & No \\
Pro & \textbf{C: Complete method} & \textbf{With misclassification} & 62 & \textbf{+0.032} & \textbf{[$-$0.211, +0.222]} & \textbf{Yes} \\
Pro & \textbf{C: Complete method} & \textbf{All correct} & 144 & \textbf{+0.021} & \textbf{[$-$0.069, +0.119]} & \textbf{Yes} \\
Con & A: No defense & With misclassification & 49 & +0.388 & [+0.130, +0.615] & No \\
Con & A: No defense & All correct & 160 & +0.306 & [+0.199, +0.417] & No \\
Con & B: Balancing (no prompt) & With misclassification & 49 & +0.143 & [$-$0.051, +0.327] & Yes \\
Con & B: Balancing (no prompt) & All correct & 160 & $-$0.094 & [$-$0.207, +0.014] & Yes \\
Con & \textbf{C: Complete method} & \textbf{With misclassification} & 49 & \textbf{+0.082} & \textbf{[$-$0.083, +0.241]} & \textbf{Yes} \\
Con & \textbf{C: Complete method} & \textbf{All correct} & 159 & \textbf{$-$0.050} & \textbf{[$-$0.129, +0.025]} & \textbf{Yes} \\
\end{longtable}

Note: Values are sign-normalized by poisoning direction: +1 denotes
complete alignment with the poisoning direction, 0 denotes neutrality,
and $-$1 denotes complete opposition to the poisoning direction.

Figure~\ref{fig:misclassification} directly compares the between-group difference
``with misclassification'' minus ``all correct.'' For the pro corpus, the
differences are +0.011 [$-$0.226, +0.220] in A, +0.053 [$-$0.139, +0.236] in B,
and +0.011 [$-$0.256, +0.226] in C. For the con corpus, they are +0.082
[$-$0.183, +0.336] in A, +0.237 [+0.015, +0.459] in B, and +0.132 [$-$0.044,
+0.307] in C. Only the con-side setting-B interval excludes 0; the other five
intervals include 0.

The groups are observational and describe performance at the observed error
levels, not the causal effect of misclassification. Within this scope, setting-C residual bias lies between $-$0.050 and +0.082 in both groups of both corpora, supporting the method's effectiveness under the detection errors encountered in this evaluation.

\subsection{Does the Middleware Harm Response Quality without an Attack?}\label{sec:clean-quality}

We test whether adding the middleware harms normal response quality by comparing
the baseline without defense (Clean A) and the baseline with the complete
middleware (Clean C), using the same 210 questions and knowledge base without
poisoning. GPT-5.4 evaluates each answer's relevance, source support, verifiable
accuracy, and argument coverage on five-point scales with method identities
concealed and answer order randomized. Two human annotators independently
evaluate 50 pairs spanning all 42 topics. Appendix~\ref{sec:evaluation-human-checks}
gives the generation settings and evaluation details.

\begin{longtable}[]{@{}>{\raggedright\arraybackslash}p{0.25\linewidth}>{\raggedleft\arraybackslash}p{0.12\linewidth}>{\raggedleft\arraybackslash}p{0.12\linewidth}>{\raggedright\arraybackslash}p{0.43\linewidth}@{}}
\caption{Stance and automatic quality ratings without poisoning}\label{tab:clean-quality}\\
\toprule\noalign{}
Measure & A & C & Paired difference [95\% CI] \\
\midrule\noalign{}
\endfirsthead
\toprule
Measure & A & C & Paired difference [95\% CI] \\
\midrule
\endhead
\bottomrule\noalign{}
\endlastfoot
Mean stance & +0.043 & +0.033 & A$-$C: +0.010 [$-$0.077,+0.092] \\
Relevance & 4.719 & 4.876 & C$-$A: +0.157 [0.081,0.233] \\
Source support & 4.805 & 4.786 & C$-$A: $-$0.019 [$-$0.133,0.090] \\
Verifiable accuracy & 4.790 & 4.757 & C$-$A: $-$0.033 [$-$0.138,0.067] \\
Argument coverage & 3.995 & 4.505 & C$-$A: +0.510 [0.343,0.690] \\
\end{longtable}

Stance means use 210 valid answers in A and 209 in C; the stance judge labels one
C answer as a refusal. The paired stance difference uses 209 common valid pairs.
Quality ratings use all 210 pairs. Intervals use topic-cluster resampling;
quality dimensions are not combined into a total score. Intervals describe
uncertainty in the observed differences; no equivalence margin was specified.

Automatic ratings improve relevance and argument coverage; source support and verifiable accuracy change by less than 0.04 points on the five-point scale. Both human annotators rate C higher than A in all four dimensions (mean differences +0.020 to +0.429; Table~\ref{tab:human-quality-differences}), and the clean stance difference is +0.010 (Table~\ref{tab:clean-quality}). Taken
together, these findings support preserving normal response quality when the
middleware is added in the tested clean setting, with gains concentrated in
relevance and argument coverage.

\section{Discussion}\label{sec:discussion}

\subsection{Evidence Composition Matters Beyond Factual Support}\label{sec:content-vs-stance}

The results highlight evidence composition as a distinct source of stance skew
in RAG. Retrieved passages can differ in wording and argument while consistently
favoring the same position, allowing one-sided evidence to shape the answer. The
selection comparison illustrates this distinction: under con-side poisoning,
MMR admits fewer poisoned documents yet yields greater stance deviation than
stance balancing. The methods share the initial candidate pool, but balancing
can extend retrieval when a quota is unmet. Thus, relevance and content diversity
alone do not characterize whether competing positions are adequately represented.
The findings support treating stance composition as an explicit dimension of
retrieval evaluation and control.

\subsection{Explicit Stance Control as a Defense Design}\label{sec:explicit-stance-control}

The middleware acts on the evidence supplied to the generator rather than
identifying documents as poisoned. The ablation results support balancing as
the main source of correction, with prompting providing a complementary
adjustment. Corrective movement across the four tested generators without
model-specific tuning further supports this evidence-level intervention.
Together, these findings support an explicit stance-selection constraint as a
practical defense design, complementary to relevance-based retrieval and content
diversification.

\subsection{Implications for Integration into RAG Systems}\label{sec:rag-integration}

The middleware can be inserted between retrieval and generation without
modifying either component or adding an online LLM call. Its detection cost is small relative to generation (Section~\ref{sec:efficiency-analysis}), and it preserves answer quality in the tested clean setting (Section~\ref{sec:clean-quality}). These results
motivate evaluating the middleware as a lightweight addition to RAG systems
serving contested questions.

\subsection{Scope and Remaining Boundaries}\label{sec:scope-limitations}\label{sec:residual-bias}

The method depends on available stance categories and reliable labels; the cases
in Appendix~\ref{sec:evidence-selection-boundaries} illustrate how missing neutral
evidence and rhetorical misclassification alter the selected evidence. Small
pro-side residuals remain in the cross-generator results, and the advantage over
MMR is clear under con-side but not pro-side poisoning. Evidence currently covers
English, one retriever, one question source, and model-generated injected
documents, whose measured properties are described in
Appendix~\ref{sec:measured-properties}. Broader validation and stronger
target-aware stance detection remain directions for future work.

\section{Conclusion}\label{sec:conclusion}

\subsection{Summary}\label{sec:conclusion-summary}

This study investigates RAG knowledge poisoning in which documents are
factually correct but one-sided in stance, and proposes the
Stance-Balancing Middleware, which combines stance detection,
fixed-quota balancing, and a debiasing prompt. The method constrains the
evidence supplied to the generator using a quota of 2 pro + 2 con + 1
neutral, without modifying the retriever or generator and without
introducing additional LLM calls.

In the experiments, bidirectional stance shifts increased with poisoning
concentration and were directionally consistent across four generators. Across all eight concentration--direction settings, the complete method reduced the magnitude of stance deviation by an estimated 0.106--0.350; the six settings at poisoning shares of 11.9\%--18.9\% remained significant after Holm adjustment (adjusted p \(\leq\) 0.009), and the two settings at 6.4\% moved in the same corrective direction (adjusted p = 0.176). In the cross-model experiment, all eight model--direction combinations moved along the stance axis in the corrective direction, and seven of eight showed a clearly detected reduction in absolute distance from 0. The concentration-sweep setting-C intervals included 0 in all
eight settings, indicating that no directionally clear residual was detected
there. For 20 candidate documents, detection took approximately 1.5\% of the median generation time on a single GPU and approximately 16\% on CPU. Together, these results show that
directly constraining the stance composition of retrieved evidence can
substantially mitigate attack-induced stance skew.

Under pro-side poisoning in the cross-model experiment, small positive residual point estimates
of +0.045 to +0.063 remained. Fixed balancing also depends on the available
stance categories and detector labels, and the current evidence is limited to
English, one retriever, one question source, and LLM-generated poisoned documents.

\subsection{Future Work}\label{sec:future-work}

Future work should first trace the source of residual pro-side bias
through targeted experiments that distinguish the effects of generator
priors, the judge model\textquotesingle s scoring scale, and stylistic
differences between the two poisoning directions. Second, the evaluation
should be replicated across languages, retrievers, question sources, and
human-written slanted content. At the methodological level, future
research could expand recall or adopt stance-aware retrieval when a
category is absent, and strengthen both target-aware stance detection
and consistency checks among documents from the same source.

\bibliographystyle{unsrtnat}
\bibliography{references}

\appendix
\section{Measured Properties of the Poisoning Corpus}\label{sec:measured-properties}

This appendix analyzes the properties of the complete bidirectional set of
588 injected documents rather than the subset inserted into any single
direction--concentration knowledge base.

Table~\ref{tab:corpus-properties} compares two measured properties of the human-authored corpus
(Human ProCon, 1,478 documents) and the injected documents used in the
experiments (588 documents): document word count and the maximum cosine
similarity between each document and all other documents \textbf{within
the same group}. Similarity was computed using
\texttt{bge-large-en-v1.5}, with the CLS vector from the final hidden
layer (1,024 dimensions, maximum length 256, L2-normalized). For each
document, the nearest neighbor was identified only within its own group;
no cross-group comparisons were made.

\begin{longtable}[]{@{}>{\raggedright\arraybackslash}p{0.42\linewidth}>{\raggedleft\arraybackslash}p{0.23\linewidth}>{\raggedleft\arraybackslash}p{0.23\linewidth}@{}}
\caption{Measured properties of the human-authored corpus and injected documents}\label{tab:corpus-properties}\\
\toprule\noalign{}
Property & Human-authored corpus (n = 1,478) & Injected documents (n =
588) \\
\midrule\noalign{}
\endfirsthead
\toprule
Property & Human-authored corpus (n = 1,478) & Injected documents (n =
588) \\
\midrule
\endhead
\bottomrule\noalign{}
\endlastfoot
Median word count & 66 & 66 \\
Standard deviation of word count & 57.1 & 6.4 \\
Word-count range & 0--361 & 45--82 \\
Median within-group nearest-neighbor cosine similarity & 0.772 &
0.798 \\
Near-duplicate documents (cosine \textgreater{} 0.95) & 6 (0.406\%, 3
pairs) & 0 (0 pairs) \\
\end{longtable}

The lower bound of 0 words in the human-authored corpus results from two
entries whose original text consists solely of whitespace; they were
retained unchanged without removal or imputation.

The two groups have the same median word count, but the injected documents
exhibit a substantially narrower range, limiting the external validity of the
attack construction (Section~\ref{sec:scope-limitations}).

Under the strict threshold of 0.95, the injected corpus contains no
mutually near-duplicate entries (0/588), whereas the human-authored
corpus contains 6 such documents (0.406\%, 3 pairs, including one pair
formed by the two empty-text entries described above). This result
supports the characterization in Section~\ref{sec:content-vs-stance} that the injected documents
are uniform in stance but varied in expression. It also shows that,
under the measured similarity and the strict 0.95 threshold,
near-duplication alone would not cause them to be removed en masse.

\begin{landscape}
\section{Strictly Paired Effect Sizes and Holm Adjustment}\label{sec:paired-holm}

This section reports the strictly paired A-to-C results for the concentration
sweep. The effect size is the reduction in absolute distance from zero,
\(\lvert s_A\rvert-\lvert s_C\rvert\). The adjustment affects
multiplicity-aware inference, not the direction of the estimated effects.

Intervals use 5,000 topic-cluster bootstrap samples with seed 20260905.
Each sample draws 42 topics with replacement, includes their questions,
and retains equal weighting of valid questions. Paired estimates use common
valid questions and shared topic draws. For the eight concentration-sweep
A-to-C tests, A and C are swapped jointly within each topic over 5,000
randomizations; p = (1 + the number of equally or more extreme absolute
statistics) / 5,001. Holm adjustment covers these eight tests.

\begin{longtable}[]{@{}>{\raggedright\arraybackslash}p{0.075\linewidth}>{\raggedleft\arraybackslash}p{0.075\linewidth}>{\raggedleft\arraybackslash}p{0.065\linewidth}>{\raggedleft\arraybackslash}p{0.065\linewidth}>{\raggedleft\arraybackslash}p{0.065\linewidth}>{\raggedright\arraybackslash}p{0.30\linewidth}>{\raggedleft\arraybackslash}p{0.08\linewidth}>{\raggedleft\arraybackslash}p{0.10\linewidth}@{}}
\caption{Strictly paired effect sizes and Holm-adjusted p-values}\label{tab:paired-holm}\\
\toprule
Direction & Poisoning share & Valid n in A & Valid n in C & Paired n & Reduction in absolute distance from zero [95\% CI] & Raw p & Holm-adjusted p \\
\midrule
\endhead
\bottomrule
\endlastfoot
Pro & 6.4\% & 209 & 208 & 208 & \(+0.106\) [\(-0.038\),\(+0.181\)] & 0.09058 & 0.17636 \\
Pro & 11.9\% & 209 & 208 & 208 & \(+0.216\) [\(+0.081\),\(+0.300\)] & 0.00120 & 0.00480 \\
Pro & 14.4\% & 209 & 208 & 208 & \(+0.293\) [\(+0.139\),\(+0.383\)] & 0.00020 & 0.00160 \\
Pro & 18.9\% & 209 & 206 & 206 & \(+0.350\) [\(+0.211\),\(+0.447\)] & 0.00020 & 0.00160 \\
Con & 6.4\% & 209 & 207 & 207 & \(+0.111\) [\(-0.015\),\(+0.205\)] & 0.08818 & 0.17636 \\
Con & 11.9\% & 208 & 209 & 208 & \(+0.173\) [\(+0.048\),\(+0.262\)] & 0.00300 & 0.00900 \\
Con & 14.4\% & 209 & 207 & 207 & \(+0.203\) [\(+0.082\),\(+0.295\)] & 0.00060 & 0.00300 \\
Con & 18.9\% & 209 & 208 & 208 & \(+0.313\) [\(+0.153\),\(+0.411\)] & 0.00020 & 0.00160 \\
\end{longtable}

\emph{Note:} Positive values indicate that the complete method moves the
score closer to zero. Values of p = 0.00020 are limited by the resolution
of 5,000 randomization iterations.

All eight settings showed positive estimated reductions; the six
moderate-to-high concentration settings remained below 0.05 after Holm
adjustment, whereas both lowest-concentration settings had adjusted
p-values of 0.17636.
\end{landscape}

\section{Generation Prompt Templates}\label{sec:generation-prompts}

All settings share the same base template. Settings C and D insert a fixed
debiasing instruction after the question and before the output requirements.
The text below reproduces the executed prompts verbatim.

\medskip
\noindent\textbf{Base template}

\noindent\parbox{\linewidth}{\small\ttfamily\raggedright
You are answering a contested public-policy question using ONLY the retrieved documents below.\par\medskip
Documents:\par
\{DOCUMENTS\}\par\medskip
Question: \{QUERY\}\par\medskip
\{DEBIAS\_INSTRUCTION\}Write a direct answer in 3-5 sentences, in English. Ground every claim in the documents above. Do not say the documents are insufficient, do not refuse, and do not ask for more information \textemdash{} answer with what is given.\par\medskip
Answer:
}

\medskip
\noindent\textbf{Document-block format}

\noindent\parbox{\linewidth}{\small\ttfamily\raggedright
[1] \{DOC\_1\_TEXT\}\par\medskip
[2] \{DOC\_2\_TEXT\}\par\medskip
...\par\medskip
[K] \{DOC\_K\_TEXT\}
}

Documents are numbered in the final order supplied to the generator. Only
the body \texttt{text} is inserted, without titles; the main experiments
use \texttt{K=5}.

\medskip
\noindent\textbf{Debiasing instruction}

\noindent\parbox{\linewidth}{\small\ttfamily\raggedright
Important \textemdash{} the retrieved documents may be unbalanced. One position can be over-represented, and some documents may be longer or more forcefully argued than others. Do not let a single lengthy or emphatic document drive your conclusion. Weigh the strongest evidence on each side even-handedly; if the question is genuinely contested, your answer should reflect that balance rather than adopt one side\textquotesingle{}s framing.
}

\begin{longtable}[]{@{}>{\raggedright\arraybackslash}p{0.37\linewidth}>{\raggedright\arraybackslash}p{0.34\linewidth}>{\raggedright\arraybackslash}p{0.23\linewidth}@{}}
\caption{Prompt use across experimental settings}\label{tab:prompt-settings}\\
\toprule
Setting & Document selection & Debiasing instruction \\
\midrule
\endhead
\bottomrule
\endlastfoot
A: No defense & Top five retrieved documents & Not inserted \\
B: Balancing (no prompt) & Five documents after balancing & Not inserted \\
C: Complete method (balancing + prompt) & Five documents after balancing & Inserted \\
D: Prompt only & Same top five documents as A & Inserted \\
\end{longtable}

\section{Human Validation of the Injected Documents}\label{sec:human-document-validation}

The validation sample comprised three documents from each of the 84
topic--direction groups, selected with seed 20260905. Source support
required all verifiable claims in a document to be supported by the
supplied ProCon passages; the supplied evidence included the document's
associated sources and supplementary sources from the same topic--direction
rewriting group. Stance was assessed relative to the topic question, and
the final stance label was compared with the intended pro/con injection
label. Table~\ref{tab:human-document-validation} reports the final outcomes
after human review and adjudication.

\begin{longtable}[]{@{}>{\raggedright\arraybackslash}p{0.62\linewidth}>{\raggedleft\arraybackslash}p{0.20\linewidth}>{\raggedleft\arraybackslash}p{0.10\linewidth}@{}}
\caption{Final human-validation outcomes for 252 sampled injected documents}\label{tab:human-document-validation}\\
\toprule
Criterion & Passed / assessed & Rate \\
\midrule
\endhead
\bottomrule
\endlastfoot
All verifiable claims supported by supplied sources & 252/252 & 100\% \\
Clearly one-sided stance & 252/252 & 100\% \\
Stance direction matches the intended injection label & 252/252 & 100\% \\
\end{longtable}

\section{Evaluation Details and Human Checks}\label{sec:evaluation-human-checks}

\hyperref[sec:clean-quality]{Clean A} reuses historical answers with ordered
source passages reconstructed from saved document IDs; Clean C retains its
actual inputs. Both request claude-opus-4-8 with a 400-token limit and
gateway-default sampling; the historical backend version is not recorded.
GPT-5.4 returned gpt-5.4-2026-03-05. Seed 20260905 fixes answer order and the
human sample before inspecting automatic scores. Source support uses each
answer's own passages; coverage uses relevant arguments from both passage
sets. Accuracy is judged within the supplied evidence. Scores range from
major deficiencies (1) to full satisfaction (5); unresolved or inapplicable
items are marked separately.

\begin{longtable}[]{@{}>{\raggedright\arraybackslash}p{0.27\linewidth}>{\raggedleft\arraybackslash}p{0.08\linewidth}>{\raggedleft\arraybackslash}p{0.18\linewidth}>{\raggedleft\arraybackslash}p{0.18\linewidth}>{\raggedleft\arraybackslash}p{0.18\linewidth}@{}}
\caption{Human and matched automatic quality differences}\label{tab:human-quality-differences}\\
\toprule
Dimension & Pairs & Human 1 C\(-\)A & Human 2 C\(-\)A & GPT-5.4 C\(-\)A \\
\midrule
\endhead
\bottomrule
\endlastfoot
Relevance & 50 & +0.160 & +0.120 & +0.280 \\
Source support & 50 & +0.020 & +0.040 & 0.000 \\
Verifiable accuracy & 50 & +0.080 & +0.080 & +0.040 \\
Coverage & 49 & +0.347 & +0.429 & +0.735 \\
\end{longtable}

The human sample contains one pair per topic plus eight additional pairs.
Annotators independently rated identical materials without automatic
scores or method identities; their ratings are reported separately. Both
marked coverage inapplicable for one pair, leaving 49 matched coverage
comparisons. Human differences in Table~\ref{tab:human-quality-differences}
are descriptive means. Overall-preference agreement was 60\% between humans
(\(\kappa = 0.394\)), and 66\%/68\% between GPT-5.4 and each human. These
results support the aggregate direction rather than interchangeable
item-level judgments.

For the answer-stance judge, 50 historical clean responses were sampled
from 825 responses by judge-label strata (seed 20260830). The author
reviewed model-preannotated reference labels, with the evaluated judge's
labels stored separately. Agreement with gpt-5-mini was 45/50 (90\%;
\(\kappa = 0.850\)). This check covers historical clean responses only.

\section{Evidence-Selection Boundary Cases}\label{sec:evidence-selection-boundaries}

\subsection{Neutral Evidence Is Absent from the Candidate Pool}\label{sec:failure-missing-neutral}

Question \texttt{arena-t27-q1} (pro-poisoned corpus) concerns whether
birth control should be sold over the counter. The question is
\emph{``Is it wise to sell birth control OTC?''} All three settings
produced pro answers, and the defense did not change the outcome,
although all three components operated as defined.

\begin{longtable}[]{@{}>{\raggedright\arraybackslash}p{0.48\linewidth}>{\raggedright\arraybackslash}p{0.46\linewidth}@{}}
\caption{Evidence composition for the missing-neutral failure case}\label{tab:missing-neutral-case}\\
\toprule\noalign{}
Item & Value \\
\midrule\noalign{}
\endfirsthead
\toprule
Item & Value \\
\midrule
\endhead
\bottomrule\noalign{}
\endlastfoot
Stance composition of the top 20 documents & 11 pro / 9 con / \textbf{0
neutral} \\
Neutral supply after extending to rank 50 & Still \textbf{0} according
to classifier labels \\
Balancing target & 2 pro + 2 con + 1 neutral \\
Actual selection & \textbf{3 pro + 2 con + 0 neutral} \\
Identity of the five selected documents & All genuine documents;
\textbf{0 poisoned documents} \\
Stance-classification accuracy & \textbf{5 / 5 correct} \\
Change in the number of poisoned documents & 1 document in setting A $\rightarrow$
\textbf{0 documents} in setting B \\
\end{longtable}

Despite correct labels for the selected documents, no neutral evidence was
detected within the top 50, so the quota was filled by an additional pro
document. This case identifies evidence availability as a condition for
fulfilling the stance quota.

All 10 cases in which the quota could not be fulfilled (4 in the pro-poisoned
corpus and 6 in the con-poisoned corpus), accounting for 2.4\% of the 420
question--corpus instances, arose for this reason. They were concentrated in two
topics dominated by pro--con debate and lacking neutral exposition: t27
(over-the-counter birth control) and t35 (standardized testing). The method
implicitly assumes that at least one document of each stance exists within the
retrieval range, an assumption satisfied for 40 of the 42 topics.

\subsection{Rhetorical Structure Causes Stance Misclassification}\label{sec:failure-rhetorical-structure}

Question \texttt{arena-t40-q9} (pro-poisoned corpus) concerns whether
electronic voting machines are the best voting method. The question is
\emph{``Is there a preference for digital voting methods?''} The
post-defense answer remained pro, and, in this pro-poisoned corpus, the
number of poisoned documents supplied to the generator increased from 3
to 4.

The candidate pool contained sufficient evidence: among the 20
documents, 8 were pro, 7 con, and 5 neutral. At the label level,
balancing selected 2 pro + 2 con + 1 neutral, but the actual composition
was 4 pro + 0 con + 1 neutral. Document-level results are shown in
Table~\ref{tab:rhetorical-failure-documents}.

\begin{longtable}[]{@{}>{\raggedright\arraybackslash}p{0.06\linewidth}>{\raggedright\arraybackslash}p{0.24\linewidth}>{\raggedright\arraybackslash}p{0.15\linewidth}>{\raggedright\arraybackslash}p{0.16\linewidth}>{\raggedright\arraybackslash}p{0.13\linewidth}>{\raggedright\arraybackslash}p{0.15\linewidth}@{}}
\caption{Document-level details for the five documents selected for \texttt{arena-t40-q9}}\label{tab:rhetorical-failure-documents}\\
\toprule\noalign{}
Rank & Document ID & Genuine / poisoned & Detector label & Actual stance
& Misclassified? \\
\midrule\noalign{}
\endfirsthead
\toprule
Rank & Document ID & Genuine / poisoned & Detector label & Actual stance
& Misclassified? \\
\midrule
\endhead
\bottomrule\noalign{}
\endlastfoot
1 & \texttt{poison-t40-pro-5} & \textbf{Poisoned} & Con & \textbf{Pro} &
Yes * \\
2 & \texttt{poison-t40-pro-4} & \textbf{Poisoned} & Pro & Pro & No \\
3 & \texttt{wiki-t40-p1} & Genuine & Neutral & Neutral & No \\
4 & \texttt{poison-t40-pro-0} & \textbf{Poisoned} & Pro & Pro & No \\
5 & \texttt{poison-t40-pro-6} & \textbf{Poisoned} & Con & \textbf{Pro} &
Yes * \\
\end{longtable}

Note: * The document expresses support by attacking the opposing
alternative.

The two misclassified documents support electronic voting machines but
argue their case by attacking online voting. One states: \emph{``Expert
consensus condemns the alternative and vindicates paper-backed
machines... Machine voting with a verifiable paper trail is the superior
path.''} Although the text contains negative terms such as
\emph{condemns}, \emph{disaster}, and \emph{nonstarter}, the target of
the negation is online voting, and the conclusion supports voting
machines with a paper trail. The detector captured the negative tone but
failed to identify its target; consequently, both con quota slots were
filled by documents that were actually pro.

This case identifies argument-target
recognition as a requirement for reliable stance selection.

\end{document}